\documentclass{llncs}
\usepackage[utf8]{inputenc}
\usepackage[style=american]{csquotes}
\usepackage[inline]{enumitem}
\usepackage[overload]{textcase}
\usepackage{booktabs}
\usepackage{graphicx}
\usepackage[style=lncs,backend=biber]{biblatex}
\usepackage{subfig}
\usepackage[tracking=true]{microtype}

\usepackage[dvipsnames]{xcolor}

\newcommand{\term}[1]{\textit{#1}}

\newcommand{\abbr}[1]{\textls[50]{\MakeTextUppercase{#1}}}

\makeatletter
\def\blfootnote{\xdef\@thefnmark{}\@footnotetext}
\makeatother

\begin{document}
\mainmatter              
\title{Privacy at Home: an Inquiry into Sensors and Robots for the Stay at Home Elderly}
\titlerunning{Privacy at Home: an Inquiry into Sensors and Robots for the Stay at Home Elderly}  
%
\author{Trenton Schulz\inst{1} \and
  Jo Herstad \inst{1} \and
Harald Holone \inst{2}}
\authorrunning{Trenton Schulz Jo Herstad Harald Holone}   
%
\tocauthor{Trenton Schulz Jo Herstad Harald Holone}
\institute{University of Oslo, Postbox 1072 Blindern, 0316 Oslo, Norway, \\
  \email{[trentonw|johe]@ifi.uio.no},
    \and
    Østfold University College, Postbox 700, 1757 Halden, Norway,
    \email{harald.holone@hiof.no}}

\maketitle              
\blfootnote{© Springer International Publishing AG, part of Springer Nature 2018\\
J. Zhou and G. Salvendy (Eds.): ITAP 2018, LNCS 10927, pp. 377–394, 2018.
https://doi.org/10.1007/978-3-319-92037-5\_28}

\begin{abstract}
The elderly in the future will use smart house technology, sensors,
and robots to stay at home longer. Privacy at home for these elderly
is important. In this exploratory paper, we examine different
understandings of privacy and use
\citeauthor{PalenUnpackingPrivacyNetworked2003}’s framework to look
at the negotiation of privacy along boundaries between a human at home, the robot, and
its sensors. We select three dilemmas: turning sensors on and off, the robot seeing through
walls, and machine learning. We discuss
these dilemmas and also discuss ways the robot can help make the
elderly more aware of privacy issues and to build trust.
\end{abstract}

\section{Introduction}\label{sec:introduction}

A popular solution to help older people stay at home longer is to use
technology. Some examples of these solutions are \term{smart home}
technology with different sensors around the house. These sensors
detect and record different types of information: if someone is in the
room, how much someone is breathing, measuring the pulse, etc. The
information collected can be helpful as
\textcite{GoonawardeneSensorDrivenDetectionSocial2017} showed by using
sensors to detect social isolation of seniors living at home.

These sensors in the house raise privacy issues for the
elderly living at home. What are the sensors recording? Further, these
sensors will typically be connected to the Internet and may send their
data to other services to aid in machine learning to help future
algorithms and robots. Who are they sending information to? Though
many people may be unaware or indifferent to the sensors, placing all
the sensors around the home may make the elderly feel they are under
constant surveillance or they don’t have any privacy at
home. An older person may feel they have gained independence at
the cost of being watched. Not to mention how
visitors to the home may feel about the surveillance.

A different solution may be to introduce a robot into the home. The
robot can carry many of these sensors and provide a mobile way of
watching over the elderly person. A robot does not eliminate the need
for sensors around the house since the robot may need stationary sensors to
navigate and perform its duties. But it may offer a better way for the
elderly person to relate to the sensors and understand the privacy
issues involved with this technology. In addition, the robots
could provide the elderly with an opportunity to negotiate their
privacy since they could have an idea where the robot is instead of sensors
that may be hidden around the house.

We are engaged in a research project investigating how a robot can
assist elderly living at home. How is the potential privacy of a
person affected by these sensors and robots in the home? What ways can
the robot carry out its functions while preserving privacy? How can
the robot help inform the elderly about privacy issues? How can we
model these interactions between sensors, robots, and people? These
are all general questions that need to be addressed. In this
exploratory paper, we look at three dilemmas: turning sensors on
and off, sensors that can see through walls, and machine learning. Our
contribution is to highlight some privacy issues that appear when a
robot is in the home of the elderly with different sensors, and make
it possible to consider them in designing future human-robot
interaction (\abbr{hri}).

In the following paper, we will provide some background about the
Internet of Things, privacy, and robots. We will use a privacy
framework to show the boundaries of negotiation between the elderly at
home, the robots and their sensors. We will present and discuss three
dilemmas before discussing future directions and concluding the paper.

\section{Background}\label{sec:background}

In the following section, we will present background on
 the Internet of Things, smart homes, privacy, and
robots. Though these concepts seem intuitive, it is useful
to examine how they are used here.

\subsection{The Internet of Things and the Smart Home}

The Internet of Things (IoT) was originally meant as an idea for
different things communicating in their own self-contained networks
using technology like \abbr{rfid} \parencite{AshtonThatInternetThings2009}. But as
devices and radios became more powerful and more energy efficient, it has
changed to include the idea of items or “things” that communicate with
each other over the Internet
\parencite{AtzoriInternetThingssurvey2010}. For example, wireless
sensor networks that are used for monitoring patients
\parencite{LeisterQualityServiceAdaptation2011} can be a form of the
IoT.  The \term{smart home} introduces concepts like
ubiquitous computing using devices from the IoT.

As more devices are added that have a network connection outside the
home, it is important to consider the privacy
issues. Though each smart home is different, they likely contain
devices and sensors that help the home do things or assist the people
at home. For example,
\textcite{NationalPublicRadioSmartAudioReport2017} looked at people
owning speakers that are connected to online services like Amazon or
Google (\term{smart speakers}) and found that people owning them tend
to change habits to incorporate the speaker more into their daily
routines. However, privacy issues have been highlighted by others
\parencite{EstesDonBuyAnyone2017} including the fact that they are
always listening for commands.

A question that is raised when discussing sensors and robots is
\term{trusting} them. Does the person trust the device to do what it
should do and preserve the person’s privacy?
\Textcite{SchulzCreatingUniversalDesigned2014} looked at issues of
creating trustworthy objects that can exist in smart homes, especially
for people with disabilities or older
people. \Textcite{BuschAllTargetingTrustworthiness2013} showed that
the different requirements could be balanced to create a more
trustworthy smart home artifact and that including people in the
process helped create that result.  The issue of trusting the device
and assuming that people in the smart home can preserve their privacy
has also been tested out in virtual and real smart home environments
\parencite{BuschBeingThereReal2014,SchulzCaseStudyUniversal2014}.
Though focusing more on the IoT in general,
\textcite{FritschInternetThingsTrust2012} discussed different
strategies one could use when interacting with different items where
one cannot determine if one can or should trust the
item.

\subsection{Privacy in a networked world}

Privacy and technology have been an issue for decades. As
\textcite[p. 195]{WarrenRightPrivacy1890} wrote:

\blockquote[][]{Instantaneous photographs and newspaper enterprise have invaded the
sacred precincts of private and domestic life; and numerous mechanical
devices threaten to make good the prediction that “what is whispered
in the closet shall be proclaimed from the house-tops.”}

Yet it can be difficult to come up with an agreed upon definition of
privacy. In performing a literature search in human-computer interaction (\abbr{hci}), computer
supported cooperative work (\abbr{cscw}), and ubiquitous computing,
\textcite{CrabtreeRepackingPrivacyNetworked2017} could not find any agreement
 on the concept of privacy. Rather they found
understandings of privacy could be divided into several groups:

\begin{description}
\item [{Control}] 
 Privacy is
understood as controlling the flow of personal information to who sees
it. This is often attributed to \textcite{WestinPrivacyFreedom1967}.
\item [{Boundary}] 
 Privacy is
understood as a set of boundaries that are negotiated by the person and
who will have access. The basis for this understanding comes from
\textcite{AltmanEnvironmentSocialBehavior1975}.
\item [{Contextual Integrity}] 

An understanding that is presented by \textcite{NissenbaumPrivacycontextualintegrity2004} as a flow of information that is appropriate based on the context. Others, like \textcite{EssPersonalDataChanging2013} have used this understanding to show that privacy is evolving beyond to individual to include the idea of protecting relationships with others.
\item [{Paradox, Trade-off, and Concern}] 

Privacy is a paradox in that some people say
one thing about privacy, but do another. Privacy can also drive
concerns about what is being done with the personal information, and
the trade-off of giving this information against its benefit.
\item [{Protective Measure}] 

Viewing entities that want access to personal
information as an attack or threat, privacy is understood as
protection against these attacks. This often focus on the technology
and ways of implementing technologies
\parencite{BellottiDesignPrivacyUbiquitous1993}. Privacy by design
\parencite{LangheinrichPrivacyDesignPrinciples2001} is also a part of
this research. Another area of examination is the leaking of
information through a \term{side channel}. That is, information leaked
as a side effect of what you are doing
\parencite{PercivalThoughtsSpectreMeltdown2018}. Side channels can be technological or social.
\end{description}

All of these understandings of privacy are useful.  In this paper, we
will use privacy as negotiation based on boundaries (i.e.,
part of the Boundary understanding from above).

Building on the privacy work done by
\textcite{AltmanEnvironmentSocialBehavior1975}, \textcite{PalenUnpackingPrivacyNetworked2003}
proposed a framework that looks at privacy in a networked
world. They identify three
boundaries for negotiating privacy:

\begin{description}
\item [{Disclosure boundary}] 

This boundary represents what you decide to
tell others (disclose) and what you keep to yourself. Examples
include: writing or speaking opinions about a subject in a public
forum, placing posters on your lawn, and wearing clothing for a sports
team.
\item [{Identity boundary}] 

This boundary represents the different roles we
have in our lives. For example, in some areas we are an employee,
other areas an enthusiast, and others a friend. Each of these roles
have different kinds of commitments on what types of information we
can and cannot share.
\item [{Temporal boundary}] 

This boundary represents persistence of
information. This includes how information is handled over time and
how building a history can reveal things about
someone. The persistence of information often means you lose control
over who and in what context information shows up at a later point in
time. For example, knowing you made a phone call to a person versus
knowing who, when, and the duration of all your phone calls.
\end{description}

There are other theories that have been built on top of
\citeauthor{AltmanEnvironmentSocialBehavior1975}’s work. For example,
\Textcite{PetronioBoundariesprivacydialectics2002} developed
Communication Privacy Management theory that creates rules for
boundaries and disclosures. This is a more complex way of examining
privacy issues, but \textcite{PalenUnpackingPrivacyNetworked2003} is
more straight forward for this paper’s explorations.

The framework has been used in contexts outside of the workplace
examples in the original paper. For example,
\textcite{HoloneNegotiatingPrivacyBoundaries2010} used
\citeauthor{PalenUnpackingPrivacyNetworked2003}’s framework to
examine privacy issues in a social mapping application for
accessibility issues. The framework highlighted the tension between
keeping information private versus making it public. It also
highlighted dangers of marking a workaround for accessibility as
passable as an individual versus marking it passable as a representative of the
handicap association since the latter may deter motivation to make a
proper solution.

\subsection{Robots and privacy}\label{sec:robot-and-privacy}

The term \term{robot} is used to describe things like algorithms,
artificial intelligence, agents that live in a program (e.g., a \term{chatbot}
on \abbr{irc} or Slack), automated vehicles, or just “something new,”
especially when it has to deal with something that replaces a
person. For this paper, we use the definition from the American
Heritage dictionary where a robot is defined as “a machine or device
that operates automatically or by remote control,”
\parencite{Robot2011}.  That is, the paper looks at the machine, its
sensors, and the software involved.

A robot has all the issues of other devices in the smart home and
more. Robots need sensors to find their place in the environment or
react to it. These sensors can gather different types of information,
such as recording an image or
audio. \Textcite{KandaHumanRobotInteractionSocial2012} describes how
many robots that interact with people (\term{social robots}) are dependent on sending the
information from the sensors to other computers in a network to
process the data and send responses back to the robot. The result is
that the robot has more computing power than it would have due to its
size or power constraints. But this transfer of information over the
network can result in breaching the privacy of people in the area
working with the robot.

Robots and privacy is a topic that is still being
researched.  Some early discussion of robots and privacy is from
\textcite{KahnJr.WhatHumanpsychological2007} and
\textcite{Feil-SeiferBenchmarksevaluatingsocially2007} who proposed
privacy as one of the benchmarks for evaluating human-robot
interaction.  \Textcite{SyrdalHeknowswhen2007} interviewed people for
a robot in a home scenario and asked what the robot should record. No
one they interviewed was completely comfortable with a robot recording
the information, but tolerated it if it was for an obvious purpose.

\Textcite{YoungAcceptableDomesticRobots2009} used social psychology to
examine models for the acceptance of \term{domestic robots} (i.e.,
robots that are in the
home, but are not necessarily communicated with socially). \Citeauthor{YoungAcceptableDomesticRobots2009} noted
that domestic robots would enter into personal spaces and deal with
privacy issues. In the end, they found several factors that affect
acceptance and perception of acceptance.

\Textcite{CaloRobotsPrivacy2010} presented an overview of the privacy
issues around surveillance and the fact that we act differently around
anthropomorphic social robots. Later,
\Textcite{CaloDronePrivacyCatalyst2011} posited that drones carrying
cameras in public areas could make it easier for citizens to recognize
the need for privacy.

A robot’s sensors and what they do may not be obvious.  A
study by \textcite{LeeUnderstandingUsersPerception2011} showed that
people were not aware of the sensing capabilities of the robot (for
example, that it could see behind itself) or a difference in what it
collected and what it processed.  This unawareness may even extend to
standard cameras. Yet \textcite{Caineeffectmonitoringcameras2012} ran
an experimental study with a camera, a stationary robot, and a mobile
robot to see how older adults changed their behavior to preserve their
privacy. \Citeauthor{Caineeffectmonitoringcameras2012} found that the
older adults exhibited the most privacy-preserving behaviors when a
camera was used and not a mobile robot.
\Textcite{Schaferspymylittle2017} discussed this lack of transparency
and other privacy (and copyright) issues. They argued that designs
need to be more obvious for the people that will be using or
interacting with the robot. Other projects
\parencite{DraperEthicalvaluessocial2017,AmirabdollahianAccompanyAcceptablerobotiCs2013}
have seen this need for respecting the privacy of an elderly person at
home.

Some of the current research on robots and privacy has focused on
robots that are operated by another person remotely and allows the
person to be present and perform tasks in the environment where the
robot is located (\term{telepresence} or \term{teleoperated robots}). The focus of
this research is on obscuring the environment from the robot
operator. \Textcite{ButlerPrivacyUtilityTradeoffRemotely2015} studied
people’s perceptions of privacy, and how well different video filters
affected the operator’s performance. Other types of filters have also
proven effective
\parencite{HubersUsingVideoManipulation2015,HubersVideoManipulationTechniques2015}.
In a different type of experiment,
\Textcite{RuebenEvaluationphysicalmarker2016} used different
interfaces for marking objects that should remain hidden to a robot’s
camera.  In another experiment,
\textcite{RuebenFramingEffectsPrivacy2017} found that informing a
person who was operating the robot was important to the person’s
privacy concerns and what the robot did in the person’s home.

Others have examined laws regarding privacy and robots.
\Textcite{PagalloRobotscloudprivacy2013} provides a summary of
the different issues with privacy, Internet connectivity, robots, and
what people can expect for
privacy. \Citeauthor{PagalloRobotscloudprivacy2013} argues for the
\abbr{EU} to examine these issues more.
\Textcite{FoschVillarongaEuropeanregulatoryframework2017a} examined
privacy issues with \term{carrier robots} (i.e., robots that carry people).
They determined that the person has more control over the carrier
robot than social robots, resulting in fewer privacy issues.

\Textcite{Eysselroleselfdisclosurehumanrobot2017} ran a study
looking at interactions between a robot and whether the robot
revealing something about itself and being likable would result in the
human to reveal something as
well. They found
that it did not matter how much the participants liked the robot, but
rather how much a participant anthropomorphized the robot.

In summary, there is a diverse and ongoing research in robots and privacy. We will take the
framework from \citeauthor{PalenUnpackingPrivacyNetworked2003} to
examine elderly negotiate their privacy at home with a robot.

\section{Examining Privacy Boundaries for the Elderly at Home}

\Citeauthor{PalenUnpackingPrivacyNetworked2003}’s framework can be used to
examine the how the elderly negotiate privacy at home.
Let’s break this down into parts: the setting and
devices—with a special look at the robot, the boundaries, and finally
interactions between the boundaries and the connections.

\subsection{The Home Setting}

\begin{figure}[htb]
\centering
\includegraphics[width=\textwidth]{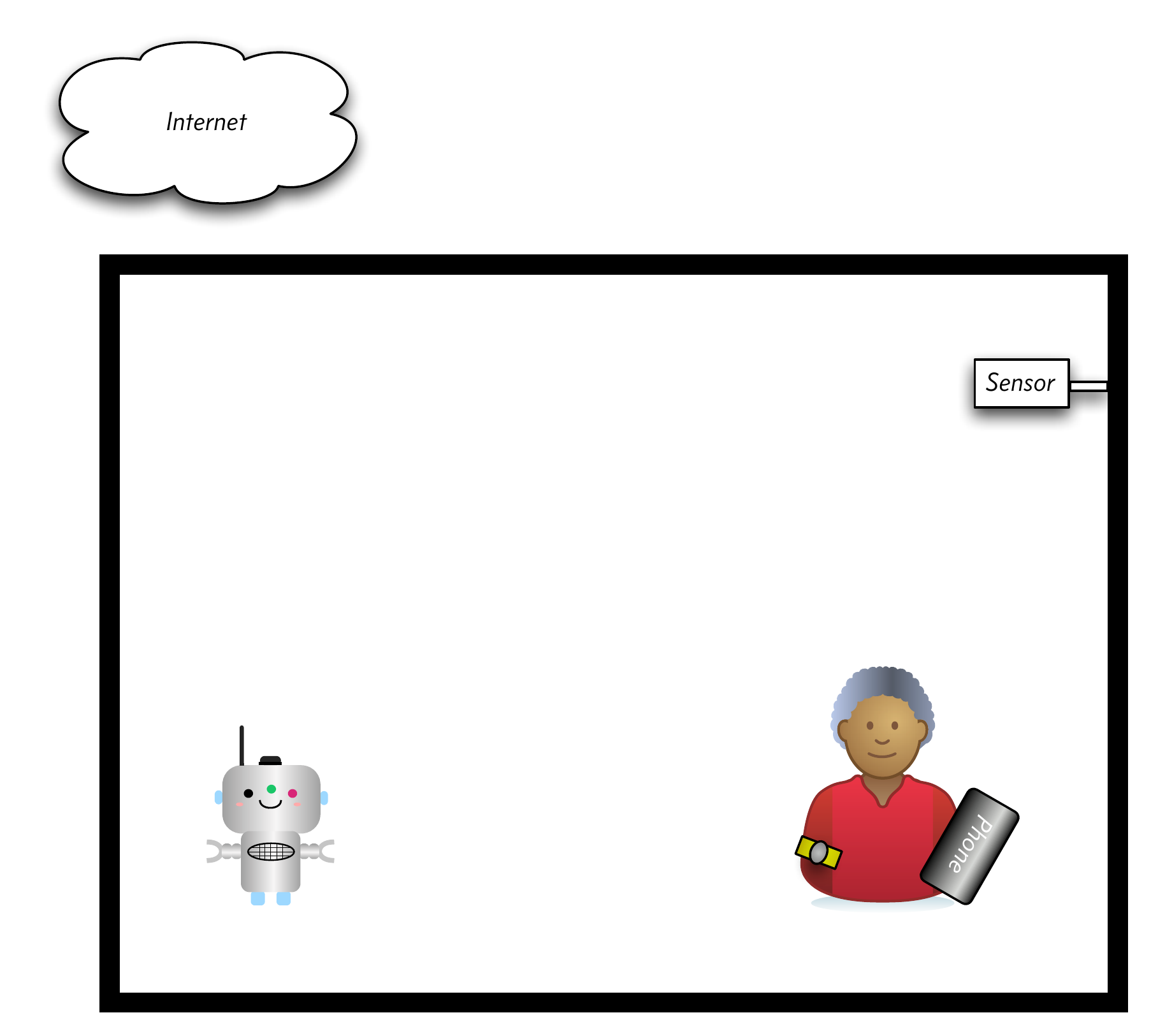}
\caption{Example set up for examining privacy issues in a smart home with a robot; the person also has a phone and a wearable device.}\label{fig:robot-model-devices}
\end{figure}

Imagine a person at home. This home has been outfitted with a
robot and sensors for helping in monitoring the person. In addition,
the person also has a wearable device and a smart phone that can
monitor the person (Fig.~\ref{fig:robot-model-devices}). The person
is elderly and there is risk of falling, the robot is not strong enough
to help the person up, but it can signal others.

It will be rare for the person to only have a robot and no other
connected devices. As mentioned in Sections~\ref{sec:introduction} and
\ref{sec:robot-and-privacy}, the robot is likely dependent on
other sensors to help it find its way. It is also likely that either
the robot or the sensors are connected to the Internet and
exchanges data to aid in its monitoring work or as input to an
algorithm to help the robot navigate around the home.  The elderly
person may also have extra devices around the house like smart
speakers, phones, wearable devices, and other sensors. The interaction
between the elderly person, the robot, its sensors, and other possible
sensors and devices is interesting, it is also complex. For this
paper, the focus will be on the elderly person, the robot, and its
sensors and connections.

\subsection{The Robot and Sensors}\label{subsec:robot-and-sensors}

\begin{figure}[htb]
\centering
\includegraphics[width=0.65\textwidth]{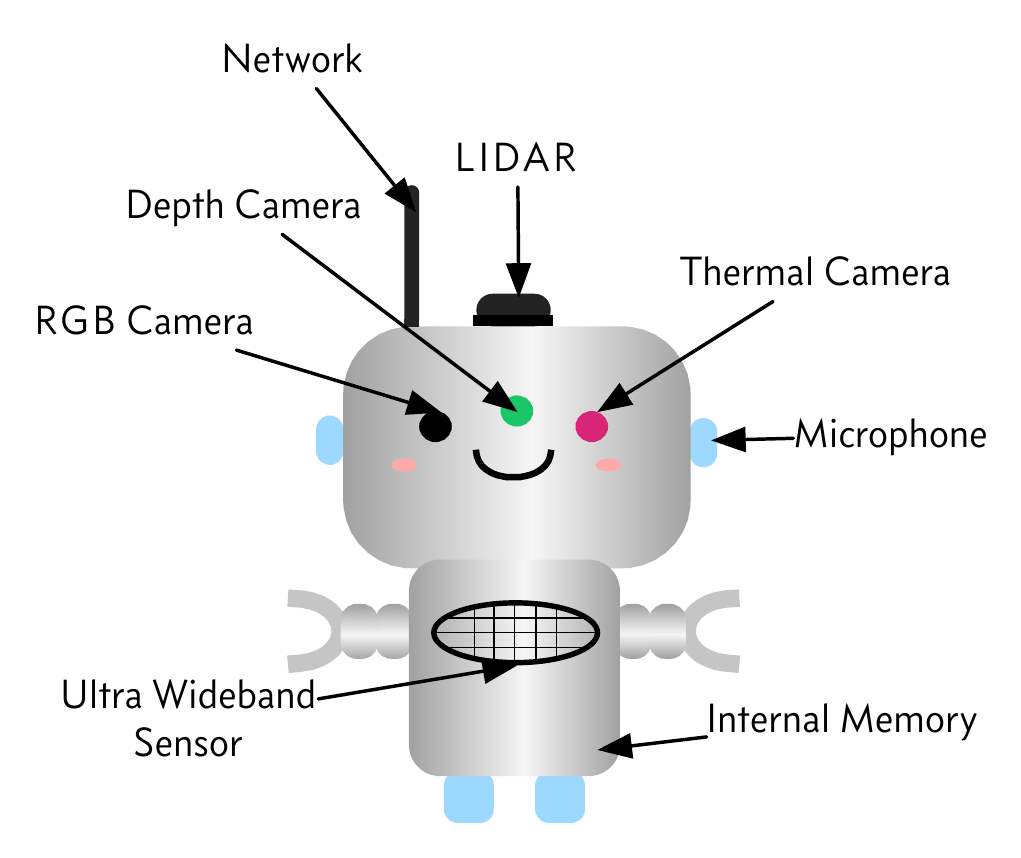}
\caption{Sensors that are on the robot in our model.}\label{fig:robot-dissection}
\end{figure}

The example robot in this situation
(Fig.~\ref{fig:robot-dissection}) needs several kinds of sensors to
perform its work, and it also has some ways of communicating. For
simplicity, the sensors are all on the robot, but many
of these sensors could also be used standalone in a smart home
environment or the sensors could be mounted remotely and relay
information back to the robot.

These sensors can be sorted by the two main senses: sight and
sound. For sound, we have the standard microphone for listening to the
person and the robot’s environment. There may be more than one
microphone to help the robot determine where the sound is coming
from. The microphone may be an input device for human-robot
interaction (i.e., the person can interact with the robot by speaking
to it). The actual “listening” may not happen on the robot; it might
be recorded by the microphone and then uploaded to another computer
for processing like is done with smart speakers.

The next two technologies use radio waves. One is an ultra-wide band
sensor as described by \textcite{TommerBodycoupledwideband2016}. These
kind of sensors can be very sensitive and can be used to detect
movement from several meters away (e.g., breathing or heart
rate). These sensors can also detect this movement through walls. They
can be thought of as special ears for detecting motion.

The second radio wave technology the robot has is a standard wireless
connection (cellular, \abbr{WI-FI}, or both) that can be used for
communicating with other networked devices (e.g., processing the sound
above). This communication can get values from other
sensors that may not be on the robot (e.g., the phone, wearable
device, and sensor in the house in
Fig.~\ref{fig:robot-model-boundaries}).

Our
example robot is fitted with several kinds of eyes. One sensor is the
Light Detection and Ranging device (or \abbr{LIDAR} for short). This
sensor works by sending out a pulsed laser beam that can be used to
measure the robot’s distance from nearby objects. \abbr{LIDAR} is
typically used to help a robot find its place in an area, and to
notice things that may have moved since the last time it was
there.

\begin{figure}[!htb]
\centering
\subfloat[Depth camera\label{fig:depth-camera}]{
\includegraphics[width=0.45\textwidth]{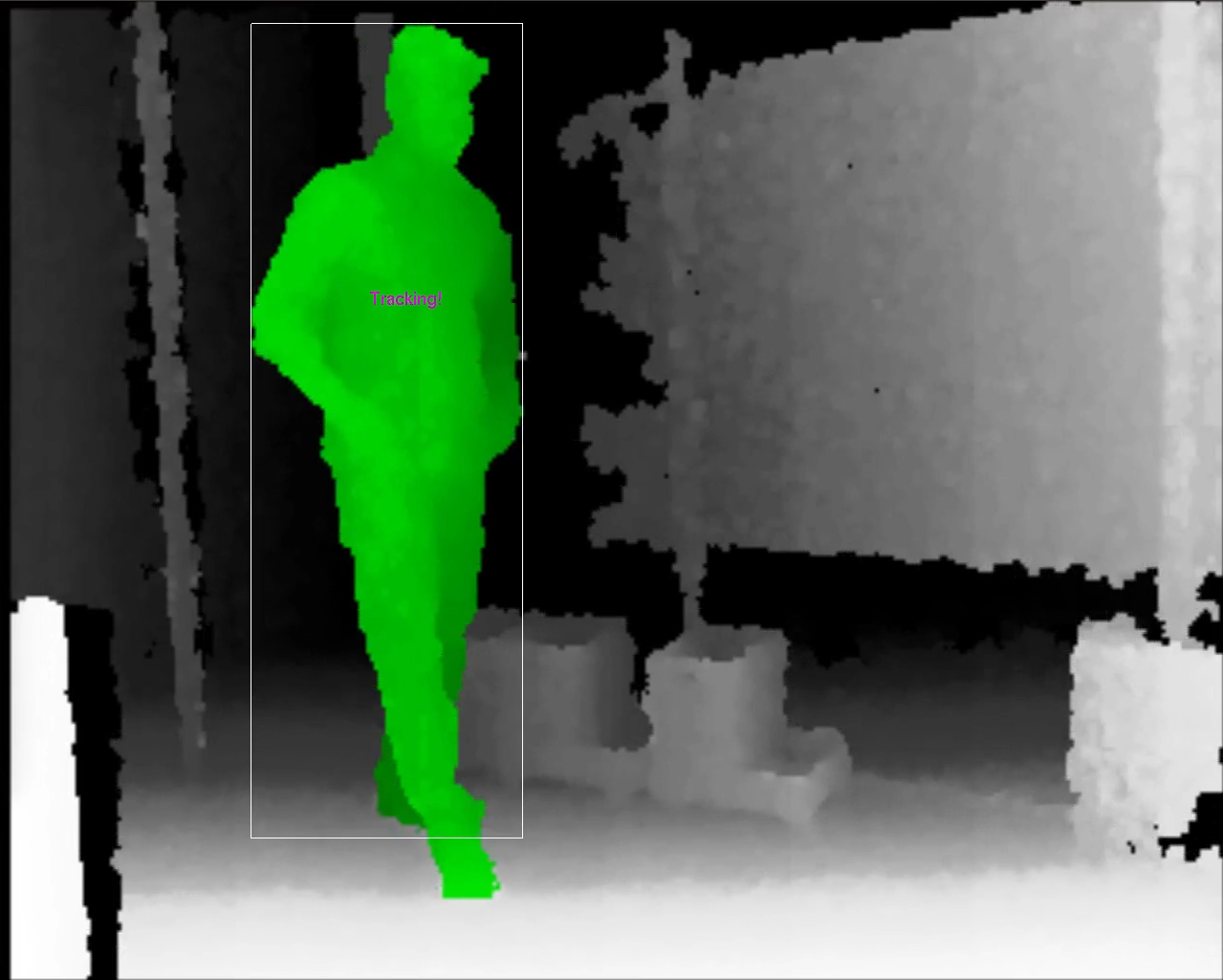}
}
\hfill
\subfloat[Thermal camera\label{fig:thermal-camera}]{
\includegraphics[width=0.45\textwidth]{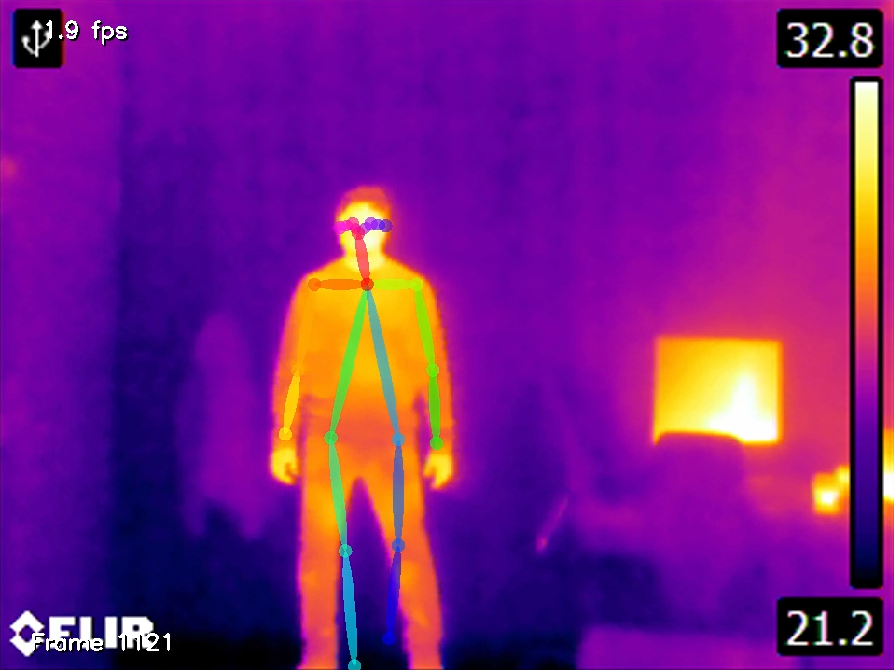}
}
\caption{Example images from a depth camera (a) and a thermal camera (b).}
\end{figure}

The example robot has three cameras. First, there is the \abbr{RGB}
(Red, Green, Blue) camera, a regular camera
that captures visible light. Second, there is a depth camera. Like
\abbr{LIDAR}, this camera can show the distance between itself and an
object. But \abbr{LIDAR} typically has a fixed height on a robot, while a
depth camera will build depths of the entire scene it can see (for
example, Fig.~\ref{fig:depth-camera}). Finally, there is the thermal
camera. This monitors infrared light to calculate the temperature of
the objects it is pointed at.

Pictures from the cameras are useful and straightforward to interpret
for people. But algorithms can analyze these images and determine
where a person is in the system and create a skeleton for the person
(Fig.~\ref{fig:thermal-camera})

Finally, there is the robot’s storage and
memory. The robot can record data from its sensors and use this data
for later use. There is a limit to how much memory and storage
a robot has, but the robot may also be linked to the Internet and the data
may be stored some place else.

\subsection{Negotiating Privacy}\label{sec:money}

\begin{figure}[htb]
\centering
\includegraphics[width=\textwidth]{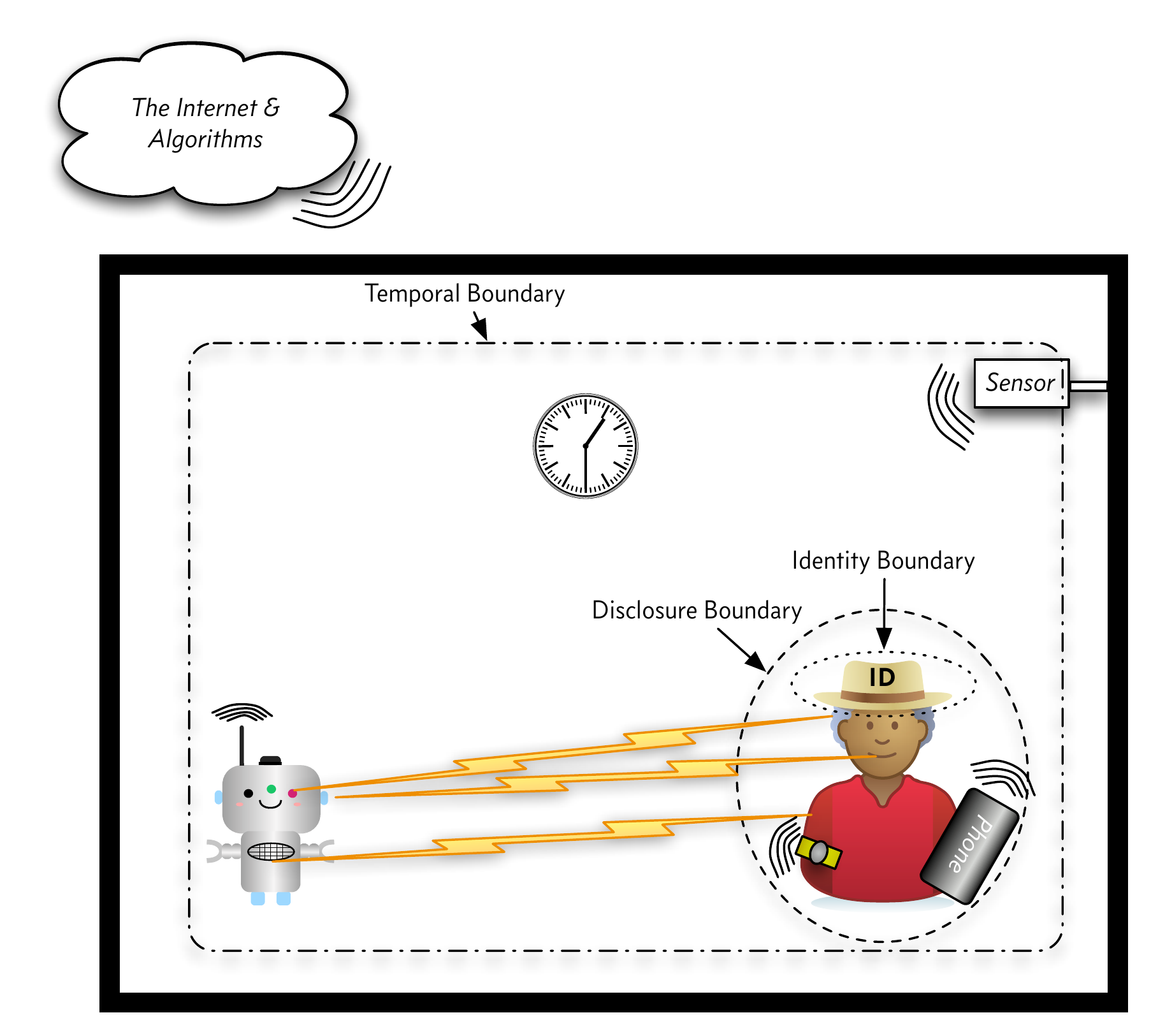}
\caption{The disclosure, identity, and temporal boundaries between the robot and the person along with the connections between the sensors.}\label{fig:robot-model-boundaries}
\end{figure}

\textcite{PalenUnpackingPrivacyNetworked2003} boundaries help show
areas where privacy is
negotiated. Fig.~\ref{fig:robot-model-boundaries} depicts boundaries
of the elderly person, the robot, and the ways that that different
conditions for negotiation. To summarize the boundaries:

\begin{description}
\item [{Disclosure Boundary}] 

Represented by the dashed line in Fig.~\ref{fig:robot-model-boundaries}. In the home
scenario, anything the elderly person says or does can reveal information about what is going on, what the person is doing, or where the person is. This
includes who is in the house and what is said. In
Fig.~\ref{fig:robot-model-boundaries}, the person negotiates the disclosure boundary when the person is sensed by the robot.

\item [{Identity Boundary}] 

Represented by the dotted line and also
represented by the \emph{ID} hat in
Fig.~\ref{fig:robot-model-boundaries}. This is the identity of the
person and is represented by a hat to emphasize that a person’s role
changes depending on the situation (for example, an elderly person may
be a grandparent, friend, or president of the gardening club).
This also shows that someone may hide other identities (here,
hats you don’t see) that only show up for certain occasions (for
example, putting on the gardening club hat for a meeting). In Fig.~\ref{fig:robot-model-boundaries},
this is blurred with the disclosure boundary, but the elderly person negotiates the identity boundary with robot by deciding if the robot can detect the hat.

\item [{Temporal Boundary}] 

Represented by dashed and dotted line going
around the clock on the wall, the robot and the person in
Fig.~\ref{fig:robot-model-boundaries}. This boundary is difficult to
visualize in a static picture, but this represents time spent together
with the robot and the person. A key factor here is also that the
robot needs to remember the information from before. This also raises
issues about why data is collected and what is to be used for. There
is a need for the elderly person to negotiate with the temporal
boundary when deciding what information the robot records and makes
use of it later.
\end{description}

There are multiple ways to negotiate privacy with the robot. This is
represented by the different (figurative) lightning bolts to and from
the different parts of the human and robot. In addition, the different
devices with the wave symbols can communicate with each other via a
wireless radio; this includes the Internet represented by the cloud
outside of the home.

Sometimes this negotiation is explicit between the robot, sensors, and
the elderly. For example, the elderly person is in the same room as a
robot. Other times, depending on how services
work, the robot may end up doing these negotiations on the elderly
person’s behalf. For example, the robot understands speech by
uploading recorded speech up to the Internet. Here, the robot has
made the negotiation decision with the disclosure boundary to
the Internet.

Negotiating on these boundaries (or the robot doing it on behalf of the
person) is \emph{not} intrinsically bad. The example robot is supposed to help with warning
about the elderly person falling. To do that on its own, it needs to
know where the person is and what the person is doing. This requires
trust from the elderly that the robot’s use of information will
be responsible.

\section{Dilemmas and Discussion}

Let’s examine some dilemmas that can come up with having a robot at
home with the elderly. These three dilemmas help illustrate some
privacy negotiations the person does with the robot.  First, each
dilemmas is presented. Next, the negotiation of privacy within this
dilemma is presented. This is followed with
some general discussion. For the first two dilemmas
(Sections~\ref{sec:dil-turn-sensors-off} and
\ref{sec:dil-see-through-walls}), only the disclosure and
identity boundaries will be presented; the third dilemma’s
(Section~\ref{sec:dil-machine-learning}) coverage of the temporal
boundary is relevant for the other dilemmas.

\subsection{Dilemma 1: Turning Sensors On and Off}\label{sec:dil-turn-sensors-off}

Having control over the sensors is an obvious way of negotiating
privacy. This means that the robot cannot detect the person while
sensors are off. The robot may not be able to do much either. The
connections would look like Fig.~\ref{fig:robot-model-devices}.

But it is not obvious how one can turn off all the controls in the
system with robots and sensors. At the same time, the robot needs the
sensors to perform its duties.  How can the elderly person easily know which
sensors are on and off to negotiate their privacy? The elderly person can ask
the robot to turn on and off the sensors, but this requires trust that
the robot will do the right thing.

\subsubsection{Privacy Negotiation}

\begin{description}
\item [{Disclosure Boundary}] 

The negotiation here is the elderly person’s desire
for privacy. By turning on or off the sensors, the elderly person is
explicitly stating a desire for privacy or willingness to share information with the
robot.
\item [{Identity Boundary}] 

An elderly person’s role may also be part of
the negotiation with turning sensors on and off. For example, the
person could be hosting visitors and doesn’t want the robot in the
way. This role of host is the reason the sensors are turned off. On
the opposite side, a person could be ill or be temporarily disabled
and need the robot on more than usual. This disabled role could also
be deduced.
\end{description}

\subsubsection{Discussion}
Turning on and off a robot’s sensors is not different from other
ways people negotiate surveillance in their everyday lives. For example,
many people cover the cameras that are embedded in the displays of
their laptops and phones. They remove these covers when they want
to use the camera, but keep them covered otherwise. Even if
the camera is on, it can only record darkness.

Another issue is the elderly person’s desire for privacy
versus the purpose the robot in the home.  What happens when the
elderly person needs the robot and cannot turn on the robot? Even if
the robot can re-activate itself, if the person is out
range of the robot, how can the robot find the person? How can
it distinguish the person having left the house versus just being
out of sensor range?

This balance between privacy and safety needs to be handled.  This
dilemma was also flagged by
\textcite{AmirabdollahianAccompanyAcceptablerobotiCs2013} in their
robot project. They suggested that the elderly person would have to
figure out (i.e., negotiate) this balance. The elderly person would need
to accept some loss of privacy in some situations. This has also been
confirmed by discussions and interviews we’ve had with elderly. Some
admit that they need help and are willing to have some sort of
monitoring for this to happen. Another point is to figure out if
turning off the sensors is recorded separately than other reasons for
turning off the sensors (for example, a system reboot).

One way of finding this balance is to use a time limit for how long
the sensors can be off—a \term{snooze}. In the situation where the
robot is snoozing and the elderly need help, it may be a matter of
waiting for the robot to wake up.  This still leaves open the need to
negotiate with all the other sensors if true privacy was desired. It
may also be difficult to actually find and turn off all these sensors
as well. Returning to the laptop example above, it is easy to cover
the web camera, but it can be difficult to cover the accompanying
microphones (if you can find them). This leads to the second dilemma.

\subsection{Dilemma 2: Seeing Through Walls}\label{sec:dil-see-through-walls}

\Textcite{Schaferspymylittle2017} put forth the idea that as long as
you cannot see the robot, then the robot cannot see
you. \Textcite{SchulzWalkingAwayRobot2018} framed this as the idea of
walking away from the robot. Of course, some elderly people may have
issues with mobility, so the robot could also have the possibility of
walking away, especially if it is asked.

This follows how we negotiate privacy with
other people. If you can put something or some distance between you
and the eyes of others (or the robot’s sensors), you probably have
some privacy.
Fig.~\ref{fig:robot-model-boundaries} shows that this walking away
model could work. If the elderly’s boundaries are out of range of the
robot’s cameras and microphones (or hidden from them), there is no
interaction with the robot. Putting up a privacy wall between the
robot and the elderly can accomplish this for cameras, and speaking
softly may also circumvent the microphones.
But if the robot has an ultra-wideband
sensor  that can
sense people through the wall (Section~\ref{subsec:robot-and-sensors}), then the idea of negotiating privacy becomes
difficult, and the walking away idea breaks down.

\subsubsection{Privacy Negotiation}
\begin{description}
\item [{Disclosure Boundary}] 

The ultra-wideband sensors \emph{eliminates} the chance for the person to negotiate presence (or lack of presence) in a room. This makes it difficult to trust the robot.
\item [{Identity Boundary}] 

The sensor may be able to pick up multiple
people in an area and roles such as host could be picked up. If the elderly person is often in a room, it may also indicate a specific role the person has in that room.
\end{description}

\subsubsection{Discussion}
The dilemma here is to balance the power of the sensors that can see
through walls versus the elderly person’s power to negotiate privacy.

There could be some negotiation on when the robot uses these
sensors. For example, if the ultra-wide band sensor lets the robot
sense through walls, perhaps the robot only uses this sensor
only when it has difficulty finding the person? The robot could also
do the equivalent of leaving the person alone once found. For example,
by leaving or going to another room.

The robot could be designed so it is obvious when this
sensor is in use. Another suggestion from
\textcite{Schaferspymylittle2017} is that robots sensors and what they
are doing should be obvious to the elderly what they do.
This could be a notification when it is using non-obvious
sensors. This might be through sound, indicator lights, or some sort
of other notification (for example like a recording light on a camera
or an “on”-indicator on a microphone).  This would help give the
elderly person an opportunity to negotiate privacy instead of
the robot overriding the process.

Instead of using the ultra-wide band sensors, one could choose
sensors that can cut down in the amount of information leaked by the
robot. For example, \abbr{LIDAR} is usually at a fixed height. As long as
it can see its obstacles, it doesn’t need to be much higher than the
base of the robot.  \Textcite{PyoMeasurementestimationindoor2013}
presented a prototype where a room contained a \abbr{LIDAR} and mirrors
mounted just above floor level. They could monitor the room for robot
navigation but not gather information about who people the person was
in the room.

Another option is to use cameras that mask identifiable information in
the images. Depth cameras and thermal cameras can show information
about how someone is moving, but do not show details of the face like
a regular camera does. They may also work better in situations where a
regular camera does not (for example, low light). Algorithms can be
trained to work with images from depth and thermal cameras as well.
For example, \textcite{KidoFalldetectiontoilet2009} used thermal
sensors to detect falls in
toilets. \Citeauthor{KidoFalldetectiontoilet2009} motivation was to
help preserve the privacy of elderly people, but still provide help if
the person had fallen. There is negotiation with the disclosure
boundary, but the amount of extra information the robot picks up is
reduced.

\subsection{Dilemma 3: Machine Learning}\label{sec:dil-machine-learning}

Many algorithms that robots use are based on machine learning.
Suppose the robot in Fig.~\ref{fig:robot-model-boundaries} watches the
elderly person, and it uses its sensors to move safely around the
house with the person. It also uses this information to determine when
the person is moving regularly or differently and may fall in the
future. This information is stored and used so that \emph{other} robots may
be further trained on noticing potential falls and moving around in a
house. This is probably accepted as a positive thing. But what about
when this data is then used by another algorithm to determine what
rates to charge someone for a service? Or perhaps as a way of training
a robot to capture people for law enforcement? This data has been used
for a different person than what was originally decided and even
though it may have been anonymized, the past actions of the person
have helped train something to do something the person may not have
been agreement with?

A similar issue is if the elderly person wants the robot to forget about
something. It is possible to delete the specific instance, but what
about if that data was used to train the machine learning algorithm?
How does pulling the specific data out change the training data or how
the algorithm developed from the previous data?

\subsubsection{Privacy Negotiation}
\begin{description}
\item [{Temporal Boundary}] 

There’s an interesting
negotiation between machine-learning algorithms and the temporal
boundary. Assuming that the robot must be \term{trained} or
\term{learn} how it will interact with the elderly through watching
the elderly and similar situations from others, the question becomes
what happens if the data is used for training in other algorithms than
the elderly person agreed to? In some ways, machine learning shows the implications of a negotiation with the temporal boundary on a large scale, going beyond the single person.
\item [{Disclosure and Identity Boundary}] 

There is no direct negotiation with the
disclosure and identity boundaries. But the negotiation on the temporal boundary to collect more
and more data to build algorithms may result in inferences that
indirectly touch on these boundaries.
\end{description}

\subsubsection{Discussion}

There is the
negotiation with the temporal boundary and the use of previous
information to help the robot perform its work. Since machine learning
is built up from data from previous encounters, it’s difficult for it
to work without having access to previous information.

There is an obvious technical solution to this problem. Just have the
robot delete the data that it has from time-to-time. This removes the
direct negotiation with the temporal boundary, but there is the
indirect negotiation through machine learning. Ideally, the data
should be removed from training data and the algorithm retaught, but
this may be impossible. Likely more data will be added to keep
improving the algorithm. So, in some ways, the original temporal data
may be gone, but it won’t be forgotten in the algorithms.

Ultimately, this may be something that must be controlled through
legal and technical solutions. This may make the temporal barrier a
more important barrier to negotiate for the elderly, but it doesn’t
solve the problem of removing the data forever. A strict negotiation
on the temporal boundary may result that the robot may never “get to
know” the elderly person in the home.

\section{Conclusion and Future Work}

This exploratory look at robots and sensors at home with the elderly
has shown that there are several privacy issues that need to be
considered. We’ve also seen that the elderly need to be aware of the
recording so they have an opportunity to negotiate their privacy with
the robots and sensors. We presented three dilemmas. Using
\citeauthor{PalenUnpackingPrivacyNetworked2003}’s framework, we could
show tensions between providing help for the elderly at home and
negotiating their privacy. There is no one-size-fits-all answer for
preserving privacy and having the elderly live safely at home. As the
dilemmas show, it’s an individual negotiation.

Beyond privacy, there’s also trust issues with the robot.  The robot
may be designed so that it is pleasing to the eye and easier to accept
in the home than a sensor in a wall. But the dilemmas
above gives reason to pause in trusting the robot: is it a helpful
sheep that watches you or a wolf in sheep’s clothing?

A robot needs to be trusted by
the elderly. This goes beyond pleasing appearance and
movement. The robot should be designed with protecting the
privacy of the elderly and the elderly must have an opportunity to
negotiate their privacy.

Though this exploratory work has found potential dilemmas, it
is also important to get experiences from actual elderly at home. We have
begun conducting interviews with the elderly living at home and are
analyzing the transcripts. We will use the insights we gain from
the analysis to develop prototype robots to help the elderly negotiate
privacy at home and provide service that they value.

\subsubsection{Acknowledgments}
{This work is partly supported by the Research Council of Norway as a part of the Multimodal Elderly Care Systems (\abbr{mecs}) project, under grant agreement~247697.} Thanks also to Md Zia Uddin for help in
providing pictures from depth and thermal cameras.

\printbibliography[]

\end{document}